\documentclass[twoside]{article}

\usepackage[accepted]{aistats2021}

\usepackage{hyperref}
\usepackage{microtype}

\usepackage[authoryear]{AVTcitations}
\usepackage{AVTcommands}
\usepackage{AVTenvironments}

\usepackage{subcaption}
\usepackage{booktabs}
\usepackage{adjustbox}
\usepackage{tikz}
\usepackage{multirow}

\DeclareMathAlphabet\bc{OMS}{cmsy}{b}{n}

\bibliography{references.bib}

\runningauthor{V. Borovitskiy, I. Azangulov, A. Terenin, P. Mostowsky, M. P. Deisenroth, and N. Durrande}

\begin{document}

\twocolumn[

\aistatstitle{Mat\'{e}rn Gaussian Processes on Graphs}
\aistatsauthor{\textbf{Viacheslav Borovitskiy}\textsuperscript{\ensuremath{*1,5}}\\\textbf{Peter Mostowsky}\textsuperscript{\ensuremath{1}} \And \textbf{Iskander Azangulov}\textsuperscript{\ensuremath{*1}}\\\textbf{Marc Peter Deisenroth}\textsuperscript{\ensuremath{3}} \And \textbf{Alexander Terenin}\textsuperscript{\ensuremath{*2}}\\\textbf{Nicolas Durrande}\textsuperscript{\ensuremath{4}}}
\aistatsauthor{}
\aistatsaddress{\textsuperscript{\ensuremath{1}}St. Petersburg State University \quad \textsuperscript{\ensuremath{2}}Imperial College London 
\\\textsuperscript{\ensuremath{3}}Centre for Artificial Intelligence, University College London \quad \textsuperscript{\ensuremath{4}}Secondmind\\\textsuperscript{\ensuremath{5}}St. Petersburg Department of Steklov Mathematical Institute of Russian Academy of Sciences}]

\begin{abstract}
Gaussian processes are a versatile framework for learning unknown functions in a manner that permits one to utilize prior information about their properties.
Although many different Gaussian process models are readily available when the input space is Euclidean, the choice is much more limited for Gaussian processes whose input space is an undirected graph.
In this work, we leverage the stochastic partial differential equation characterization of Mat\'{e}rn Gaussian processes---a widely-used model class in the Euclidean setting---to study their analog for undirected graphs.
We show that the resulting Gaussian processes inherit various attractive properties of their Euclidean and Riemannian analogs and provide techniques that allow them to be trained using standard methods, such as inducing points.
This enables graph Mat\'{e}rn Gaussian processes to be employed in mini-batch and non-conjugate settings, thereby making them more accessible to practitioners and easier to deploy within larger learning frameworks.
\end{abstract}

\section{Introduction}
Gaussian process (GP) models have become ubiquitous in machine learning, and have been shown to be a data efficient approach in a wide variety of applications \cite{rasmussen06}.
Key elements behind the success of GP models include their ability to assess and propagate uncertainty, as well as encode different kinds of prior information about the function they seek to approximate.
For example, by choosing different covariance kernels, one can encode different degrees of differentiability, or specific patterns, such as periodicity and symmetry.
Although the input and output spaces of GP models are typically subsets of $\R$ or $\R^d$, this is by no means a restriction of the GP framework, and it is possible to define models for other types of input or output spaces \cite{lindgren11,mallasto2018,borovitskiy2020}.

\begin{table}[b!]
\vspace*{-3.575ex}
\footnotesize\urlstyle{same}\textsuperscript{\ensuremath{*}}Equal contribution. Code available at: \textsc{\url{http://github.com/spbu-math-cs/Graph-Gaussian-Processes}}. Correspondence to: \href{mailto:viacheslav.borovitskiy@gmail.com}{\textsc{viacheslav.borovitskiy@gmail.com}}.
\vspace*{3.425ex}
\end{table}

In many applications, such as predicting street congestion within a road network, using kernels based on the Euclidean distance between two locations in the city does not make much sense.
In particular, locations that are spatially close may have different traffic patterns, for instance, if two nearby roads are disconnected or if traffic is present only in one direction of travel.
Here, it is much more natural for the model to account directly for a distance based on the graph structure.
In this work, we study GPs whose inputs or outputs are indexed by the vertices of an undirected graph, where each edge between adjacent nodes is assigned a positive~weight.

Gaussian Markov random fields (GMRF)~\cite{rue2005} provide a canonical framework for such settings.
A GMRF builds a graph GP by introducing a Markov structure on the graph's vertices, and results in models and algorithms that are computationally efficient.
These constructions are well-defined and effective, but they require Markovian assumptions, which limit model flexibility.
Although it may be tempting to replace the Euclidean distance, which can typically be found in the expression of stationary kernels, by the graph distance, this typically does not result in a well-defined covariance kernel \cite{feragen15}.\footnotemark

As a consequence, working with GPs on graphs requires one to define bespoke kernels, which, for graphs with finite sets of nodes, can be viewed as parameterized structured covariance matrices that encode dependence between vertices.
A few covariance structures dedicated to graph GPs have been explored in the literature, such as the diffusion kernel or random walk kernels \cite{kondor2002,vishwanathan2010}, but the available choices are limited compared to typical Euclidean input spaces and this results in impaired modeling abilities.

\footnotetext{Approaches of this type can still be employed, provided one introduces additional restrictions on the class of graphs and/or on allowable ranges for parameters such as length scale. See \textcite{anderes2020} for an example.}

In this work, we study graph analogs of kernels from the Mat\'{e}rn family, which are among the most commonly used kernels for Euclidean input spaces~\cite{rasmussen06,stein1999interpolation}.
These can be used as GP input covariances or GP output cross-covariances.
Our approach is related to~\textcite{whittle63,lindgren11,borovitskiy2020}, where GPs with Mat\'ern kernels are defined on Euclidean and Riemannian manifolds via their stochastic partial differential equation (SPDE) representation.
For the graph Mat\'{e}rn GPs with integer smoothness parameters we obtain sparse precision matrices that can be exploited for improving the computational speed.
For example, they can benefit from the well-established GMRF framework~\cite{rue2005} or from recent advances in non-conjugate GP inference on graphs~\cite{pmlr-v89-durrande19a}.
As an alternative, we present a Fourier feature approach to building Mat\'{e}rn kernels on graphs with its own set of advantages, such as hyperparameter optimization without incurring the cost of computing a matrix inverse at each optimization step.

We also discuss important properties of the graph Mat\'{e}rn kernels, such as their convergence to the Euclidean and Riemannian Mat\'ern kernels when the graph becomes more and more dense.
In particular, this allows graph Mat\'{e}rn kernels to be used as a substitute for manifold Mat\'{e}rn kernels \cite{borovitskiy2020} when the manifold in question is obtained as intermediate output of an upstream manifold learning algorithm in the form of a graph.

\section{Gaussian processes}

Let $X$ be a set.
A random function $f : X \-> \R$ is a Gaussian process $f \~[GP](\mu,k)$ with mean function $\mu(\cdot)$ and kernel $k(\cdot, \cdot)$ if, for any finite set of points $\v{x}\in X^n$, the random vector $f(\v{x})$ is multivariate Gaussian with mean vector $\v\mu = \mu(\v{x})$ and covariance matrix $\m{K}_{\v{x}\v{x}} = k(\v{x},\v{x})$.
Without loss of generality, we assume that the prior mean $\mu$ is zero.

For a given set of training data $(x_i,y_i)$, we define the model $y_i = f(x_i) + \eps_i$ where $f \~[GP](0,k)$ and $\eps_i \~[N](0,\sigma^2)$.
The posterior of $f$ given the observations is another GP. Its conditional mean and covariance are
\[
\mu_{\given \v{y}}(\.) &= \m{K}_{\boldsymbol{\cdot}\v{x}} (\m{K}_{\v{x}\v{x}} + \sigma^2\m{I})^{-1}\v{y}
\\
k_{\given \v{y}}(\.,\.') &= k(\.,\.') - \m{K}_{\boldsymbol{\cdot}\v{x}} (\m{K}_{\v{x}\v{x}} + \sigma^2\m{I})^{-1} \m{K}_{\v{x}\boldsymbol{\.}'}
\]
which uniquely characterize the posterior distribution \cite{rasmussen06}. 
Following \textcite{wilson20}, posterior sample paths can be written as 
\[
\label{eqn:pathwise}
f(\cdot)\given\v{y}=f(\cdot)\!+\!\m{K}_{\boldsymbol{\cdot}\v{x}} (\m{K}_{\v{x}\v{x}}\!+\!\sigma^2\m{I})^{-1} (\v{y}\!-\!f({\v{x}})\!-\!\v\eps)
\]
where $(\.)$ is an arbitrary set of locations.

\subsection{The Mat\'{e}rn kernel}

When $X = \R^d$ and $\tau = x - x^\prime$, Mat\'ern kernels are defined as
\[
\label{eqn:matern-kernel-eucl}
k_{\nu}(x,x') = \sigma^2 \frac{2^{1-\nu}}{\Gamma(\nu)} \bigg( &\sqrt{2\nu} \frac{\norm{\tau}}{\kappa} \bigg)^\nu
 K_\nu \del{\sqrt{2\nu} \frac{\norm{\tau}}{\kappa}}
\]
where $K_\nu$ is the modified Bessel function of the second kind \cite{gradshteyn14}.
The parameters $\sigma^2$, $\kappa$ and $\nu$ are positive scalars that have a natural interpretation: $\sigma^2$ is the variance of the GP, the length-scale $\kappa$ controls how distances are measured in the input space, and $\nu$ determines mean-square differentiability of the GP \cite{rasmussen06}.
As $\nu\->\infty$, the Mat\'{e}rn kernel converges to the widely-used squared exponential kernel 
\[
\label{eqn:rbf-kernel-eucl}
k_{\infty}(x,x') = \sigma^2 \exp\del[3]{-\frac{\norm{\tau}^2}{2 \kappa^2}}
.
\]
Euclidean Mat\'{e}rn kernels are known to possess favorable asymptotic properties in the large-data regime---see \textcite{stein2010, kaufman2013} for details.

In our setting, an important property of Mat\'ern kernels is their connection to stochastic partial differential equations (SPDEs).
\textcite{whittle63} has shown that Mat\'{e}rn GPs on $X = \R^d$ satisfies the SPDE
\[
\label{eqn:spde-matern}
\del{\frac{2 \nu}{\kappa^2} - \lap}^{\frac{\nu}{2} + \frac{d}{4}}f = \c{W}
\]
for $\nu < \infty$, where $\lap$ is the Laplacian and $\c{W}$ is Gaussian white noise \cite{lifshits12} re-normalized by a certain constant---see \textcite{lindgren11} or \textcite{borovitskiy2020} for details.
Similarly, the limiting squared exponential GP satisfies
\[
\label{eqn:spde-rbf}
e^{-\frac{\kappa^2}{4} \Delta} f = \c{W}
\]
where $e^{-\frac{\kappa^2}{4} \Delta}$ is the (rescaled) heat semigroup \cite{evans10, grigoryan2009}.
These equations have been studied as a means to extend Mat\'{e}rn Gaussian processes to Riemannian manifolds, such as the sphere and torus \cite{lindgren11,borovitskiy2020}. 
Since these spaces can be discretized to form a graph, we explore the relationship between these settings in the sequel.

\subsection{Gaussian processes on graphs}

A number of approaches have been proposed to define GPs over a weighted undirected graph $G = (V, E)$.
The aim is to define a GP indexed by the vertices $V$, which reflects the notion of \emph{closeness} induced by the edges $E$ and their associated weights.
This has been studied by a variety of authors, including \textcite{kondor2002,rue2005}, and others, in areas such as GMRFs and diffusion kernels.
A number of variations, such as introducing dependence on attributes contained in nodes \cite{ng2018}, are possible.
We will interpret the ideas presented in this work from multiple viewpoints in order to synthesize these different perspectives.

A number of authors, such as \textcite{venkitaraman2020, zhi2020} have studied graph-structured multi-output Gaussian processes $f : \R \-> \R^{|V|}$, where $|V|$ is the number of nodes for which the outputs dependencies should reflect the graph structure. 
Although such settings may appear different from the scope outlined above, the problem can be cast into the proposed graph GP framework by constructing a Gaussian process $f: \R \x V \-> \R$ through partial function application.

It is also possible to consider Gaussian processes $f : \c{G} \-> \R$ where $\c{G}$ is an appropriately defined \emph{space of graphs}. 
Here, each individual input to the GP is an \emph{entire graph} rather than just a node on a fixed graph.
This setting departs significantly from the preceding ones, and we do not study it in this work: a recent survey on the topic is given by \textcite{kriege2020}.

\section{Mat\'{e}rn GPs on graphs}

We now define the \emph{Mat\'{e}rn} family of Gaussian processes on graphs by generalizing their SPDE characterization to the graph setting.
This will entail introducing appropriate notions for the left-hand-side and right-hand-side of equations \eqref{eqn:spde-matern} and \eqref{eqn:spde-rbf}.
Note that since a graph is a finite set, such a Gaussian process can be viewed as a multivariate Gaussian whose indices are the graph's nodes. 
Throughout this work, we refer to these as Gaussian processes to emphasize that their covariance reflects the structure of the graph.

Let $G$ be a weighted undirected graph whose weights are non-negative---for an unweighted graph, assume all weights are equal to one.
Denote its adjacency matrix by $\m{W}$, its diagonal degree matrix by $\m{D}$ with $\m{D}_{i i} = \sum_{j} W_{i j}$, and define the \emph{graph Laplacian} by 
\[
\m\lap = \m{D} - \m{W}
.
\]
The graph Laplacian is a symmetric, positive semi-definite matrix, which we view as a linear operator acting on a $|V|$-dimensional real space.
Note that this operator should be viewed as an analog of $-\lap$ in the Euclidean or Riemannian setting.\footnote{This sign convention ensures the positive semi-definiteness of $\m\lap$ and is equivalent to the analyst's (rather than geometer's) convention for studying the continuous Laplacian.} 
Notwithstanding the different sign convention, the graph and Euclidean Laplacian are intimately linked. 
For example, the Laplacian of a graph given by a grid corresponds to the finite-difference approximation of the Euclidean Laplacian, and diffusions on graphs involve the graph Laplacian in the same way diffusion in continuous spaces involve the classical Laplacian.
If $G$ is connected, then $\m\lap$ has a nullspace of dimension $1$, following the case of compact manifolds \cite{smola2003}.
Since $\m\lap$ is symmetric positive semi-definite, it admits an eigenvalue decomposition $\m\lap = \m{U}\m\Lambda\m{U}^T$ where $\m\Lambda$ is diagonal with non-negative entries and $\m{U}$ is orthogonal.

To construct graph-theoretic analogs of the differential operators in \eqref{eqn:spde-matern} and \eqref{eqn:spde-rbf}, we introduce a notion of \emph{functional calculus} for $\m\lap$.
Let $\Phi: \R \-> \R$ be a function.
For the diagonal matrix $\m\Lambda$, let $\Phi(\m\Lambda)$ be a diagonal matrix defined by applying $\Phi$ to the diagonal of $\Lambda$ element-wise.
Define the matrix $\Phi(\m\lap)$ to be
\[ \label{eqn:Phi-matrix}
\Phi(\m\lap) = \m{U} \Phi(\m\Lambda) \m{U}^T.
\]
Although such a definition may seem arbitrary, this generalization of $\Phi$ to square matrices can be interpreted intuitively as plugging $\m\lap$ into the Taylor expansion of $\Phi$, which immediately boils down to \eqref{eqn:Phi-matrix} due to the orthogonality of $\m{U}$. 
Taking $\Phi$ to be one of
\[ \label{eqn:Phi_def}
\Phi(\lambda) &= \del{\frac{2 \nu}{\kappa^2} + \lambda}^{\frac{\nu}{2}}
&
\Phi(\lambda) &= e^{\frac{\kappa^2}{4} \lambda}
\]
gives the operators on the left-hand side of \eqref{eqn:spde-matern} and \eqref{eqn:spde-rbf}, respectively.\footnote{We use ${\Phi(\lambda) = \del[1]{\frac{2 \nu}{\kappa^2} + \lambda}^\frac{\nu}{2}}$ and ${\Phi(\lambda) = e^{\frac{\kappa^2}{4} \lambda}}$ instead of ${\Phi(\lambda) = \del[1]{\frac{2 \nu}{\kappa^2} - \lambda}^\frac{\nu}{2}}$ and ${\Phi(\lambda) = e^{-\frac{\kappa^2}{4} \lambda}}$ because of the different sign convention for the 
graph Laplacian.}
Note that the term $d/4$ present in the Euclidean and Riemannian cases to ensure regularity is not needed here.
This will result in a slightly different scaling for graph Mat\'{e}rn kernels, which for Markovian cases are indexed by integers rather than the half-integers.

\begin{figure*}[t]
    \centering
    \begin{subfigure}[b]{0.24\textwidth}
    \includegraphics[width=\textwidth]{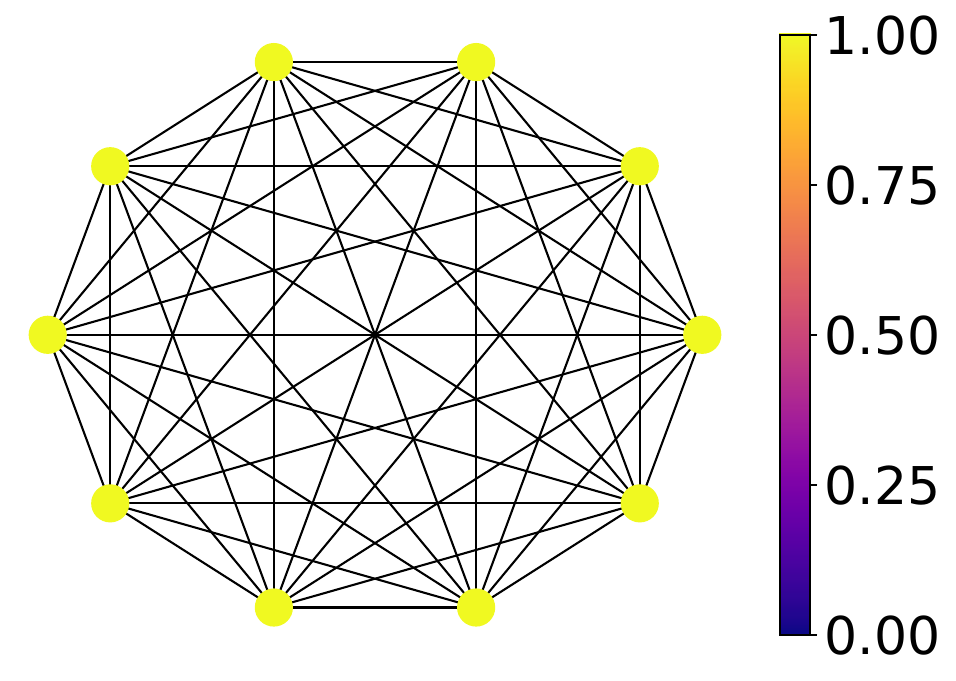}
    \end{subfigure}
    \begin{subfigure}[b]{0.24\textwidth}
    \includegraphics[width=\textwidth]{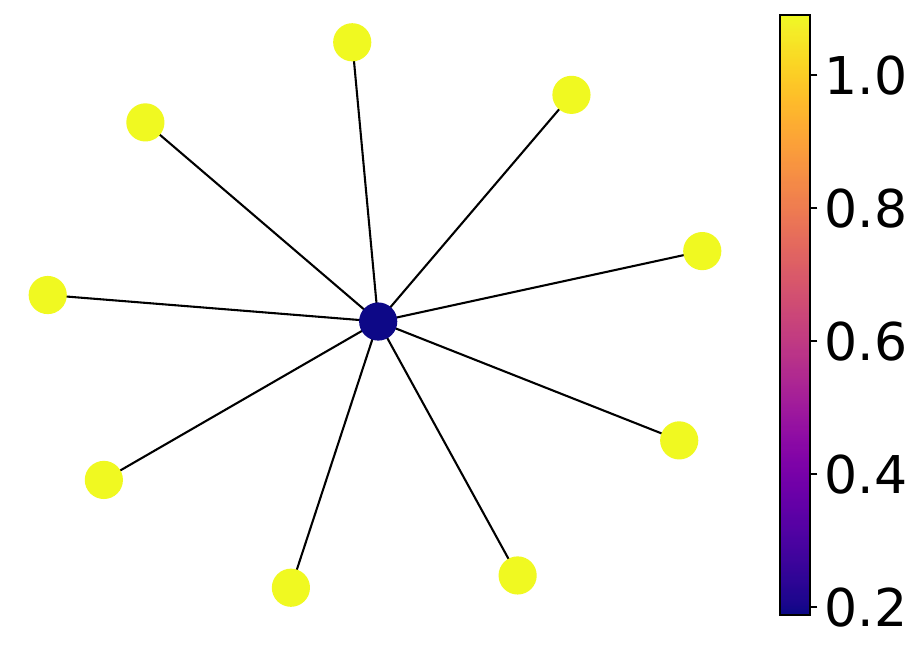}
    \end{subfigure}
    \begin{subfigure}[b]{0.24\textwidth}
    \includegraphics[width=\textwidth]{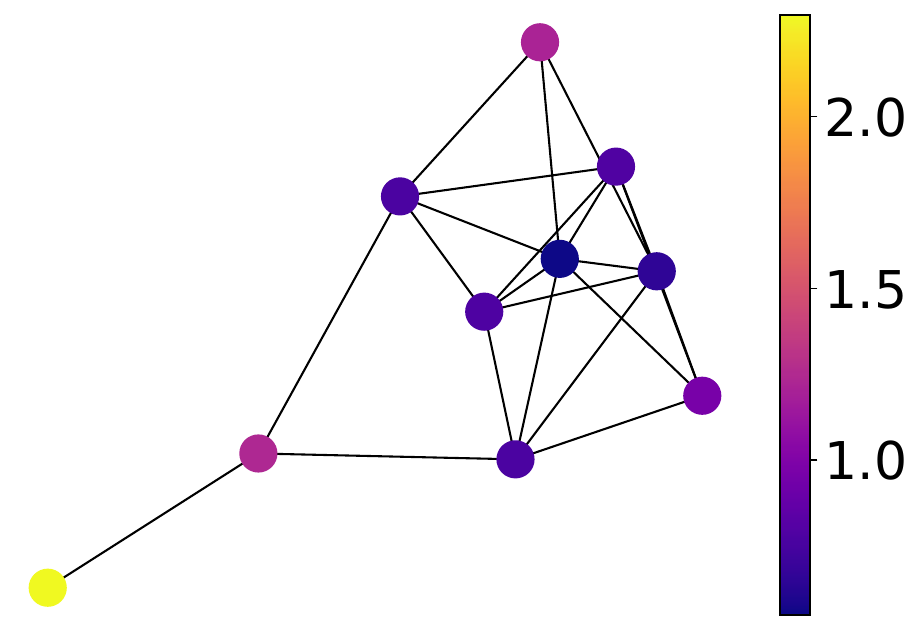}
    \end{subfigure}
    \begin{subfigure}[b]{0.24\textwidth}
    \includegraphics[width=\textwidth]{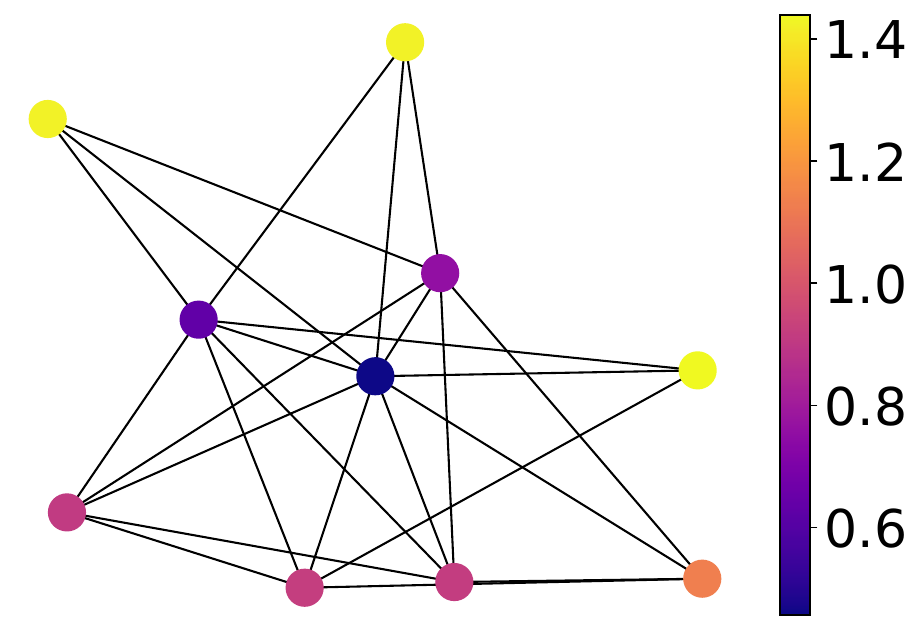}
    \end{subfigure}
    \caption{Here we illustrate prior variance for a complete graph, a star graph, and two randomly generated graphs. 
    For the complete graph,
    the variance is uniform because of its symmetry (reordering the nodes always gives the same graph). In the star graph, the center node has much lower variance, this mirrors the behavior of random walk for which the return time is lower for the center node than for the other nodes. The random graphs illustrate the idea that variance depends on graph structure rather than on the degrees of individual nodes.}
    \label{fig:variance}
\end{figure*}

Replacing Gaussian white noise process with a standard Gaussian $\bc{W} \~[N](\v{0},\m{I})$ gives the equations
\[
\del{\frac{2 \nu}{\kappa^2} + \m\Delta}^\frac{\nu}{2} \v{f} &= \bc{W},
&
e^{\frac{\kappa^2}{4} \m\Delta} \v{f} &= \bc{W},
\]
where we associate the vector of coefficients $\v{f}$ with the function $f : V \-> \R$, which gives the graph Gaussian process of interest.
This gives 
\[
\v{f} &\~[N]\del{\v{0}, \del{{\textstyle\frac{2 \nu}{\kappa^2}} + \m\Delta}^{-\nu}},
&
\v{f} &\~[N]\del{\v{0}, e^{-\frac{\kappa^2}{2} \m\Delta}}
\]
as \emph{graph Mat\'{e}rn} and \emph{graph squared exponential} Gaussian processes, respectively.
Note that the former covariance is \emph{not} obtained by adding $\frac{2\nu}{\kappa^2}$ to the entries of the matrix $\m\lap$, but rather by adding this number to its \emph{eigenvalues}, in the sense of Equation \eqref{eqn:Phi-matrix}.
By construction their covariance matrices are positive semi-definite, and we refer to them as the \emph{graph Mat\'{e}rn} and \emph{graph diffusion} kernels, respectively.
These possess a number of key properties, which we enumerate below.

\paragraph{Sparsity.} 
For sparse graphs, $\m\Delta$ is sparse.
Hence, for sufficiently small integers $\nu$ and most graphs, the precision matrices $\del{\frac{2 \nu}{\kappa^2} + \m\Delta}^\nu$ are sparse.
This can be exploited to significantly accelerate their computational efficiency, and is utilized extensively by prior work on GMRFs \cite{rue2005}.

\paragraph{Non-uniform variance.}

The prior variance for the introduced kernels varies along vertices.
This behavior is similar to the variance of the Mat\'{e}rn kernel on certain non-homogeneous Riemannian manifolds \cite{borovitskiy2020}.
Figure \ref{fig:variance} illustrates the prior  variance of the graph Mat\'{e}rn kernel on a set of graphs.
It can be seen that the kernel's variance is not simply a function of degree and depends in a complex manner on the graph in question.

\textcite{urry2013} study a similar phenomenon in the  context of \emph{random walk kernels}. They show that the variance is determined by the return time of a certain random walk defined on the graph.

\paragraph{Use with symmetric normalized graph Laplacian.}

The above kernels are defined in terms of the graph Laplacian $\m\lap$.
In some applications, it might be preferable to instead work with the \emph{symmetric normalized graph Laplacian} $\m{D}^{-1/2}\m\lap\m{D}^{-1/2}$.
Doing so in above expressions yields
\[
\v{f} &\~[N]\del{\v{0}, \del{{\textstyle\frac{2 \nu}{\kappa^2}} + \m{D}^{-1/2}\m\lap\m{D}^{-1/2}}^{-\nu}}
\\
\v{f} &\~[N]\del{\v{0}, e^{-\frac{\kappa^2}{2} \m{D}^{-1/2}\m\lap\m{D}^{-1/2}}}
\]
which we call the \emph{symmetric normalized graph Mat\'{e}rn GP} and the \emph{symmetric normalized graph squared exponential GP}, respectively.
Whether one should use the symmetric normalized or the standard Laplacian generally depends on the particular application---see \textcite[412]{von2007} for a discussion in the related setting of spectral clustering.

\paragraph{Connection with the graph diffusion equation.}
The graph diffusion kernel $e^{-\frac{\kappa^2}{2} \m\lap}$ is the Green's function of the graph diffusion equation.
That is, if $\v\phi: [0, \infty) \x V \to \R$ solves the differential equation
\[ \label{eqn:graph_diffusion}
&\od{\v\phi}{t} + \m\lap \v\phi = 0
&
& \left.\v\phi\right|_{t = 0} = \v{v}
\]
then $\left.\v\phi\right|_{t = \tau} = e^{- \tau \m\Delta} \v{v}$.
This equation describes heat transfer along the graph. 
If $\m\lap$ is replaced with the symmetric normalized graph Laplacian $\m{D}^{-1/2} \m\lap \m{D}^{-1/2}$, then the value $\left.\v\phi\right|_{t = \tau}$ can be interpreted as the unnormalized density of a continuous-time random walk which moves along the graph.

\paragraph{Limits and connection with random walks.}

\begin{figure*}[t]
    \centering
    \includegraphics[width=\textwidth]{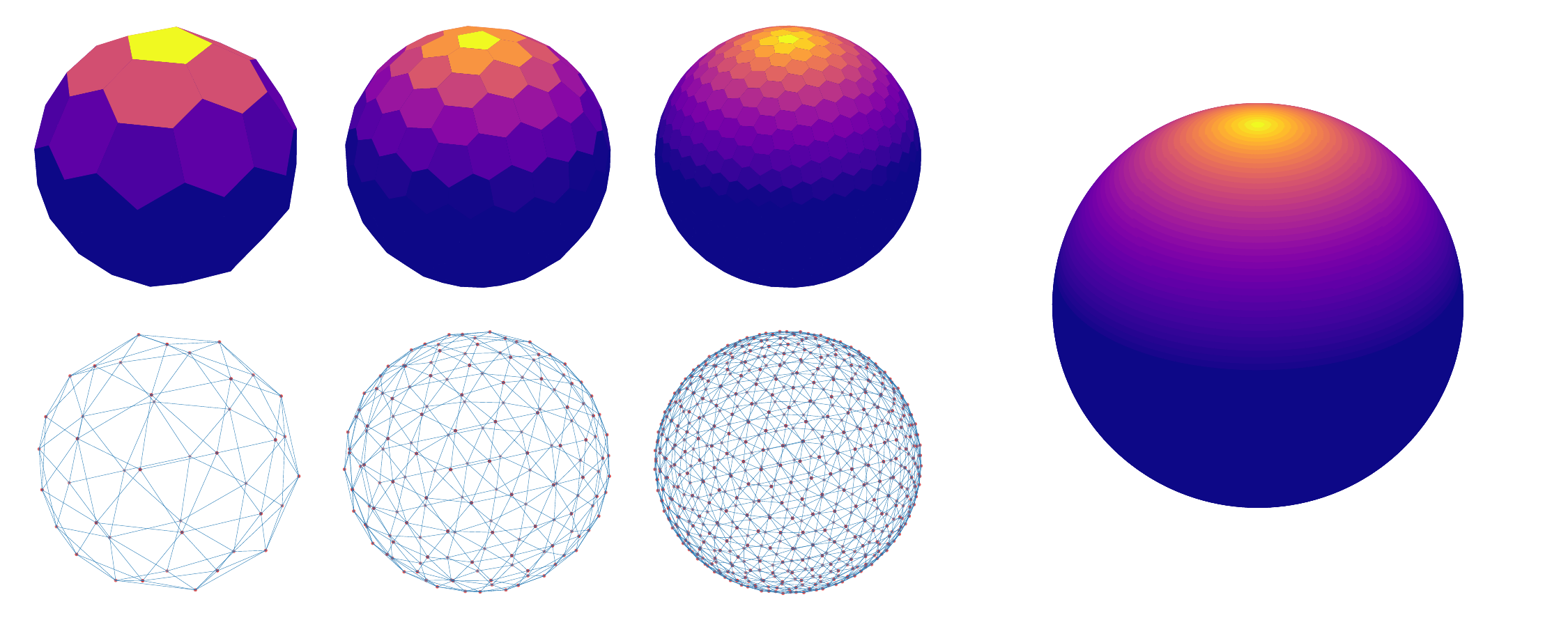}
    \caption{The Mat\'{e}rn kernel $k_{3/2}(x, \cdot)$ defined on increasingly fine triangular meshes (three left columns),
    and Mat\'{e}rn kernel $k_{1/2}(x, \cdot)$ on the sphere (rightmost). 
    The true kernel is computed using 512 analytically known Laplace-Beltrami eigenpairs via the technique in \textcite{borovitskiy2020}.
    The discrepancy in the smoothness parameter is due to a different parameterization of the kernel, which
    accounts for dimensionality of the Riemannian manifold.
    The point $x$ is chosen to correspond to the north pole. 
    For graphs defined by the meshes, kernel values are projected onto the sphere via piecewise-constant interpolation based on spherical Voronoi diagram induced by the graph nodes. 
    The graph kernel converges to the true kernel on the right. }
    \label{fig:fem_convergence}
\end{figure*}

We detail two different ways in which the graph diffusion kernel arises as a limit of a sequence of kernels.

First, mirroring the Euclidean case, the graph Mat\'{e}rn kernel converges to the graph diffusion kernel as $\nu\->\infty$, when scaled appropriately.
To see this, note that the eigenvectors of both kernels are identical, and consider the eigenvalues 
\[
(2 \nu / \kappa^2)^{\nu}
\del{\frac{2 \nu}{\kappa^2} + \lambda}^{-\nu}
\->^{\nu\->\infty}
e^{-\frac{\kappa^2}{2} \lambda}
.
\]
This implies that the corresponding matrices converge, and hence that the Gaussian processes converge in distribution.

Second, the graph diffusion kernel arises as a limit of the \emph{random walk kernel} of \textcite{smola2003}.
This kernel is defined as 
\[
\del[1]{\m{I} - (1-\alpha)\m{D}^{-1/2} \m\lap \m{D}^{-1/2}}^p
,
\]
which arises by symmetrizing the $p$-step transition matrix of a lazy random walk on the graph.
This kernel looks superficially similar to the Mat\'{e}rn kernel defined previously, but contains a number of important differences.
In particular, the power $p$ is positive rather than negative, the laziness parameter $\alpha$ is restricted to be in $[0,1)$, and the Laplacian is subtracted rather than added.
Nonetheless, this kernel also converges to the diffusion kernel, in the limit $p \->\infty$ with $\alpha = 1-\frac{\kappa^2}{2p}$ where $\kappa$ is a fixed constant.
Specifically,
\[
\begin{gathered}
\del[1]{\m{I} - (1-\alpha)\m{D}^{-1/2} \m\lap \m{D}^{-1/2}}^p 
\\
\qquad\qquad\->^{p\->\infty} e^{-\frac{\kappa^2}{2} \m{D}^{-1/2} \m\lap \m{D}^{-1/2}}
\end{gathered}
\]
which yields the normalized graph diffusion kernel.

\paragraph{Relationship with Mat\'{e}rn Gaussian processes on Riemannian manifolds.}

One of the most common ways that linear (S)PDEs in the Euclidean and Riemannian settings are solved is via the \emph{finite-element method}.
Roughly speaking, this technique expresses the (S)PDE solution using a truncated basis expansion, whose coefficients are obtained by solving a sparse linear system.
A typical choice for the finite element basis are \emph{piecewise polynomial} functions constructed using the nodes and edges of a graph, which corresponds to a mesh discretization of the manifold.

By their construction, the basis weights of a finite-element-discretized Riemannian Mat\'{e}rn Gaussian process define a GMRF approximation to the Mat\'{e}rn Gaussian process.
The precise form of the GMRF approximation obtained depends on the choice of finite element space.
Under appropriate regularity conditions, \textcite{lindgren11} have proven convergence of the finite element discretizations to the solution of the true SPDE solution.

Not all kernels that arise this way are graph Mat\'{e}rn kernels.
Fortunately, a number of other convergence results \cite{belkin2007, burago2014} are available.
One can show under appropriate regularity conditions and proper weighting of edges that the eigenvalues and eigenvectors of the graph Laplacian converge to the eigenvalues and eigenfunctions of the Laplace--Beltrami operator.
Using these ideas, \textcite{sanz2020} provide a theoretical framework for studying how graph Mat\'{e}rn kernels converge to their Riemannian counterparts.
This is illustrated in Figure~\ref{fig:fem_convergence}, which shows convergence of a graph Mat\'{e}rn kernel to its Riemannian limit for a sphere.

Graph Mat\'{e}rn kernels can also be used with graphs that are obtained as intermediate steps of manifold learning algorithms. 
This enables them to be applied to data that is concentrated on an unknown manifold in a way that captures its geometry---a common situation in practice.
Provided the manifold learning algorithm is well-posed and possesses appropriate guarantees, the resulting graph kernels will converge to the manifold Mat\'{e}rn kernels studied by \textcite{borovitskiy2020}.

\paragraph{Use as an output cross-covariance.}

The above section describes Mat\'{e}rn Gaussian processes to model functions $f : V \-> \R$.
We now briefly recall how to lift this construction to functions $\v{f} : \R^d \-> \R^{|V|}$, where the output cross-covariance is induced by the graph $G$.
The simplest way to do so is to apply the \emph{intrinsic coregionalization model} \cite{alvarez2011, goovaerts1997}.
Here, one views a multi-output function $\v{f}(x)$ as a single-output function with an additional input $f(x, i)$ such that $\v{f}(x)_i = f(x, i)$.
One then uses a separable kernel \[k((x, i), (x', i')) = k_{\R^d}(x, x') k_{G}(i, i'),\] where $k_{G}(i, i')$ is the graph kernel.
This defines model that respects the graph structure over the output space.

\subsection{Scalable Training}

Here we discuss how to train Mat\'{e}rn Gaussian processes in a scalable manner.
The techniques we present fall in one of two broad areas: (1) graph Fourier feature approximations, and (2) efficient computation via sparse linear algebra.
All of the techniques described here apply even if the graph is too large to store its full covariance matrix in memory.

Graph Fourier feature methods approximate the kernel matrix using a truncated eigenvalue expansion.
Here, we first obtain the $\ell$ smallest eigenvalues and eigenvectors of $\m\lap$ using a routine designed for efficiently computing the eigenvalues of sparse matrices, such as the Lanczos algorithm.
Since the mappings 
\[
\lambda &\|>\del{\frac{2 \nu}{\kappa^2} + \lambda}^{-\nu}
&
\lambda &\|> e^{-\frac{\kappa^2}{4} \lambda}
\]
are decreasing functions on $\R^+$, we thus obtain the $\ell$ largest eigenvalues of the Mat\'{e}rn or diffusion kernels.
Once this is obtained, scalable training proceeds by relating the approximate GP with a Bayesian linear regression model, mirroring Euclidean Fourier features \cite{rahimi08}.
Here, reduced computational costs come from the orthogonality of features, i.e. the eigenvectors of the kernel matrix. This mirrors ideas in \textcite{dutordoir2020,solin2020,borovitskiy2020} where Gaussian processes are specified using Karhunen--Lo\'{e}ve type decompositions---for further details, see Appendix~\ref{apdx:theory}.
The main difficulty with this approach is that it can exhibit \emph{variance starvation} issues, which causes approximation quality to deteriorate in the large-data regime \cite{wang2018}.

A different way to train graph GPs scalably is by restricting $\nu$ to small integer values, for which in most graphs the inverse kernel matrices, termed \emph{precision} matrices
\[
\del{\frac{2 \nu}{\kappa^2} + \m\Delta}^\nu
\]
are sparse.
The prior is then a \emph{Gaussian Markov random field} for which a variety of scalable training procedures are already available off-the-shelf \cite{rue2005}.
This approach generally avoids variance starvation, at a cost often no worse than that of graph Fourier features.
The main limitation is that the computational gains diminish when $\nu$ increases, and that the numerical routines needed for scalable training may not be well-supported by standard Gaussian process packages. 
This support is rapidly improving: as shown by \textcite{pmlr-v89-durrande19a, adam2020doubly}, it is possible to leverage certain sparse computations in automatic differentiation frameworks, such as TensorFlow, to accelerate variational GP models and their sparse counterparts.

\begin{figure*}[t]
    \centering
    \begin{subfigure}[b]{0.47\textwidth}
        \begin{tikzpicture}[
                every node/.style={anchor=north east,inner sep=0pt},
                x=1mm, y=1mm,
              ]   
             \node (fig1) at (0,0)
               {%
                    \adjustbox{trim={.125\width} {.18\height} {0.1\width} {.19\height},clip,width=\linewidth,height=0.25\textheight}%
                    {\includegraphics{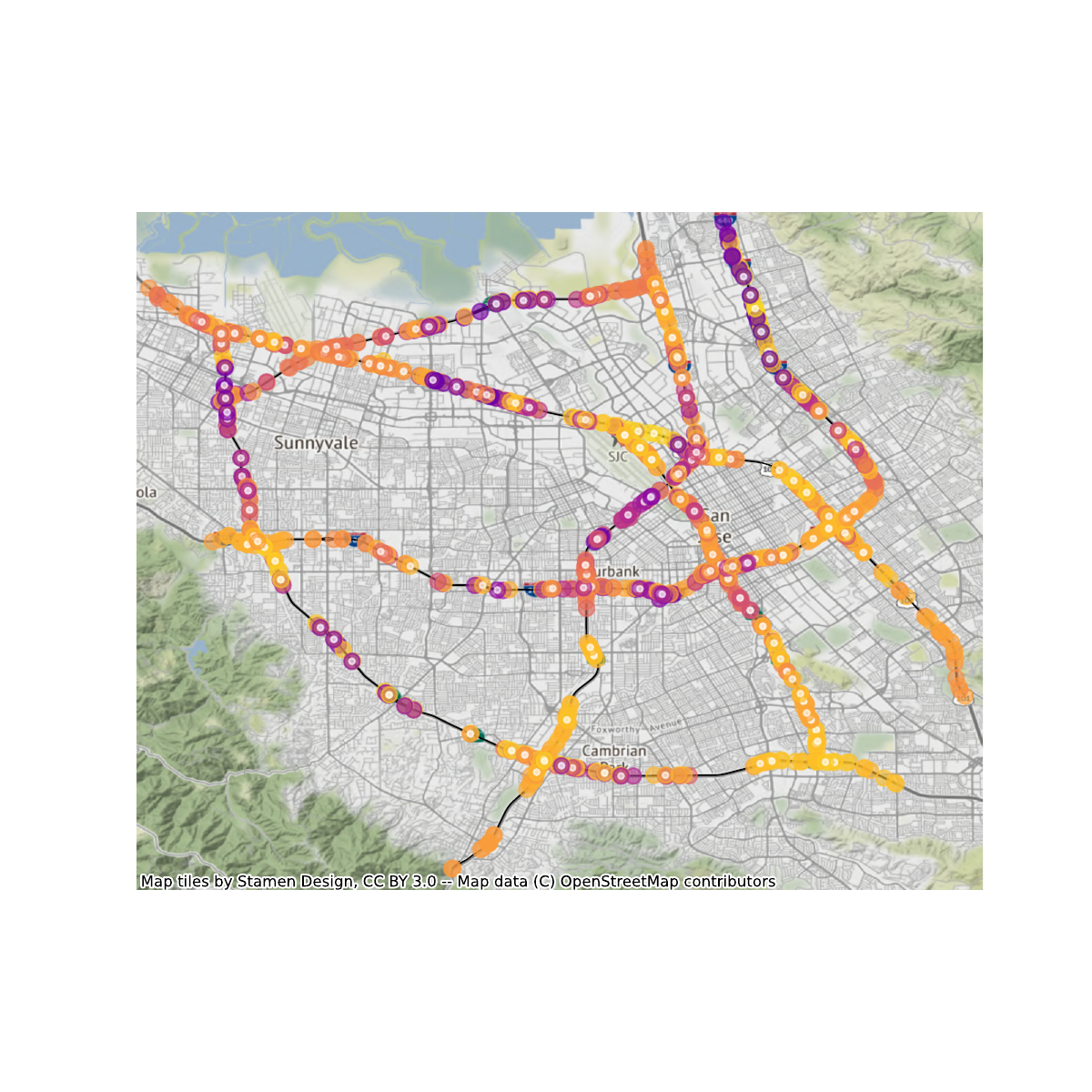}}
               };
             \node (fig2) at (0,0)
               {
                    \adjustbox{trim={0} {0} {.1\width} {.08\height},clip,width=0.6\linewidth,height=0.14\textheight}%
                    {\includegraphics{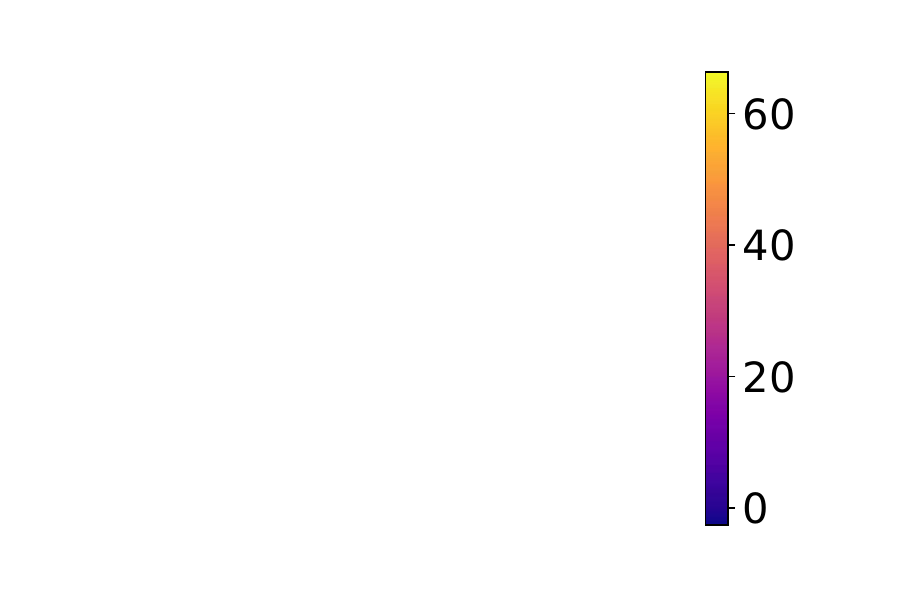}}
               };  
        \end{tikzpicture}
        \caption{Mean}
    \end{subfigure}
    \hspace{0.04\textwidth}
    \begin{subfigure}[b]{0.47\textwidth}
        \begin{tikzpicture}[      
                every node/.style={anchor=north east,inner sep=0pt},
                x=1mm, y=1mm,
              ]   
             \node (fig1) at (0,0)
               {%
                    \adjustbox{trim={.125\width} {.18\height} {0.1\width} {.19\height},clip,width=\linewidth,height=0.25\textheight}%
                    {\includegraphics{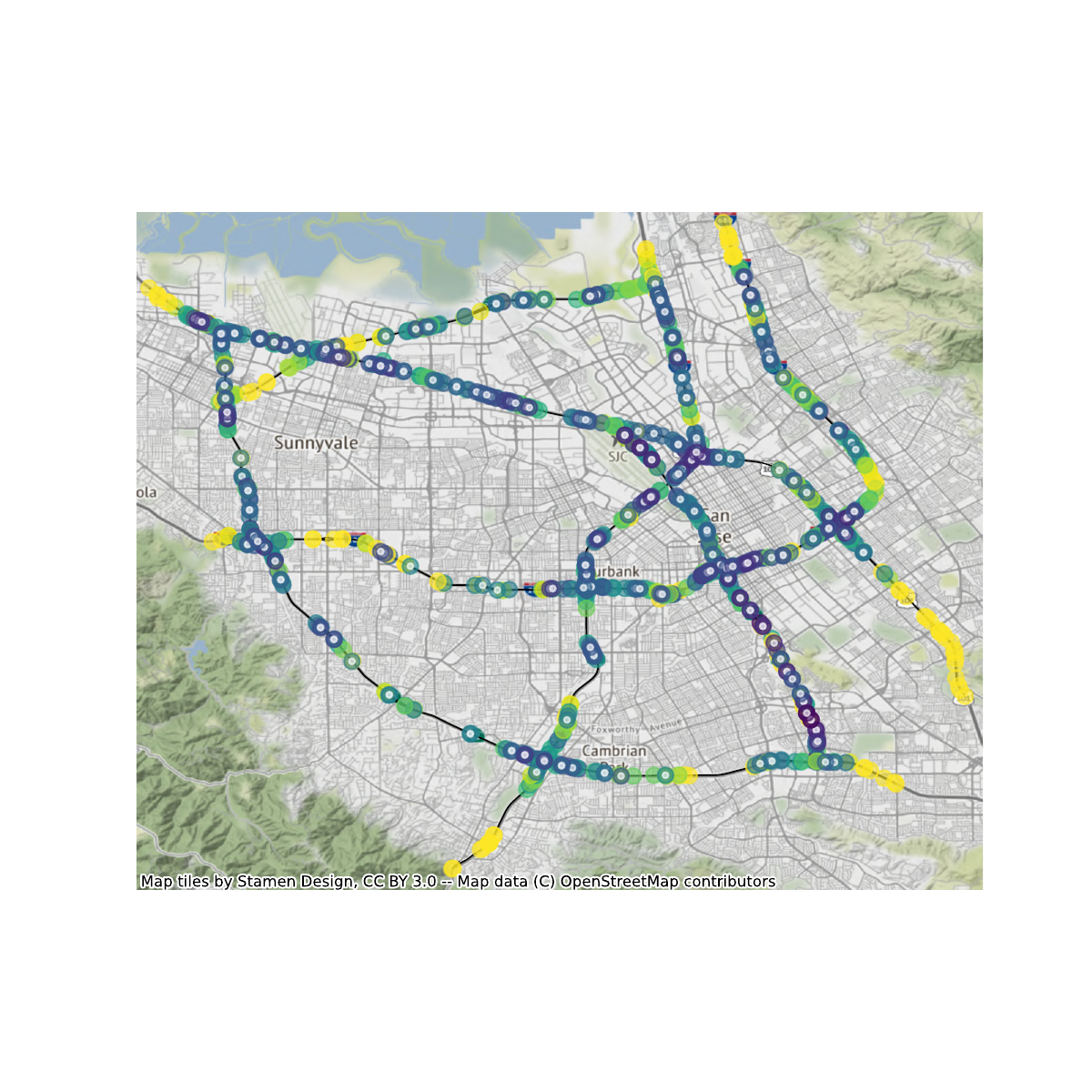}}
               };
             \node (fig2) at (0,0)
               {
                    \adjustbox{trim={0} {0} {.1\width} {.08\height},clip,width=0.6\linewidth,height=0.14\textheight}%
                    {\includegraphics{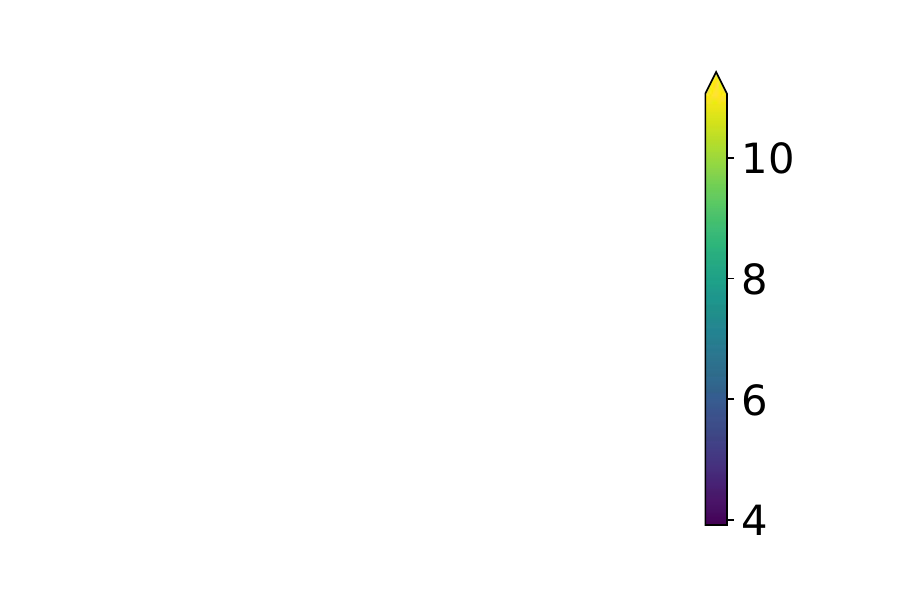}}
               };  
        \end{tikzpicture}
        \caption{Standard deviation}
    \end{subfigure}
\caption{Traffic flow speed (mph) interpolation over a graph of San Jose highways. Colored circles represent graph nodes, black lines represent edges and colored circles with a white point in the middle represent training data. The standard deviation color bar is clipped: the $10\%$ most variable points are painted in yellow.}
\label{fig:traffic_global}
\end{figure*}

\begin{figure}[t]
    \centering
    \begin{subfigure}[b]{0.475\linewidth}
        \adjustbox{trim={.125\width} {.20\height} {0.1\width} {.22\height},clip,width=\linewidth,height=0.125\textheight}%
        {\includegraphics{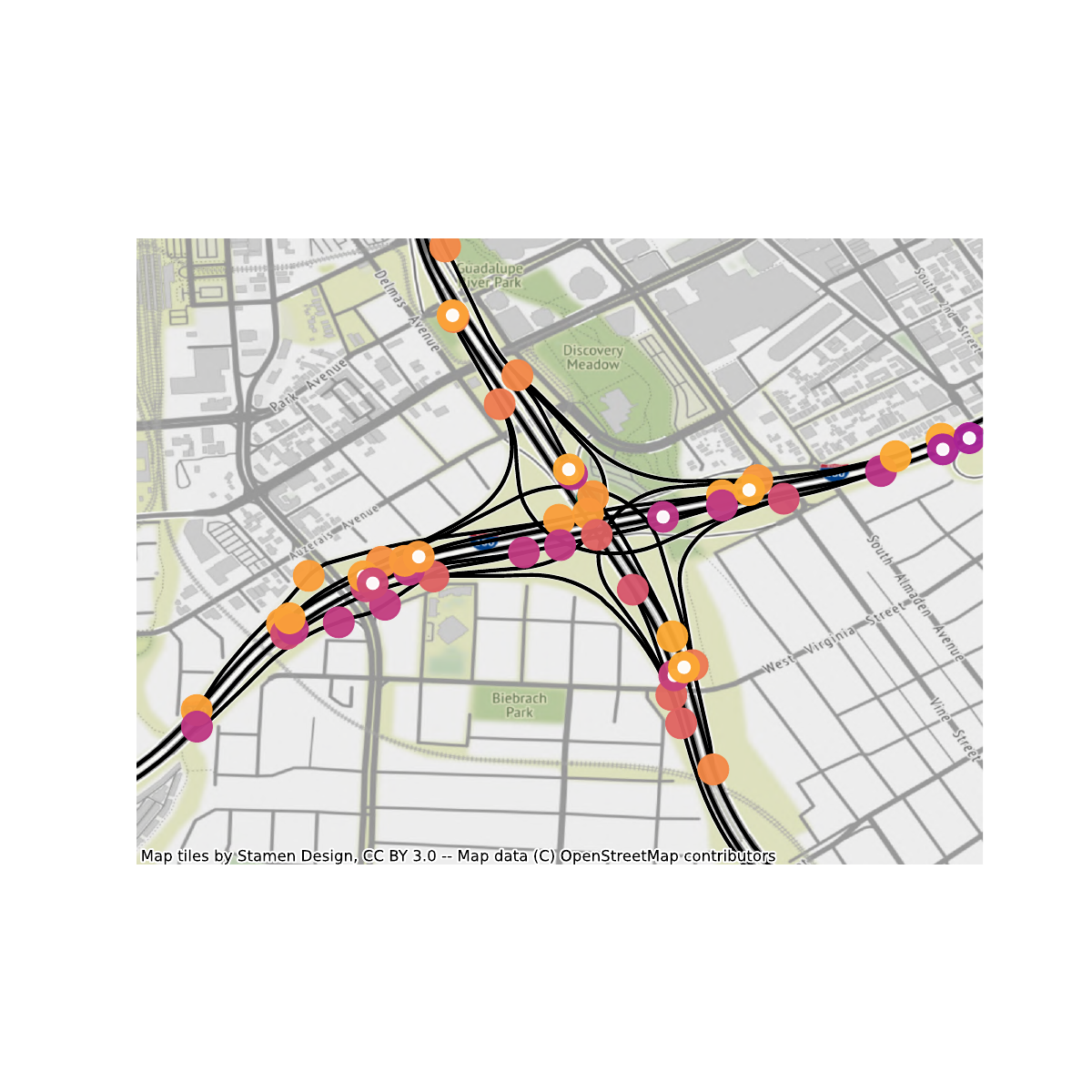}}        
        \caption{Mean}
    \end{subfigure}
    \hspace{0.01\linewidth}
    \begin{subfigure}[b]{0.475\linewidth}
        \adjustbox{trim={.125\width} {.20\height} {0.1\width} {.22\height},clip,width=\linewidth,height=0.125\textheight}%
        {\includegraphics{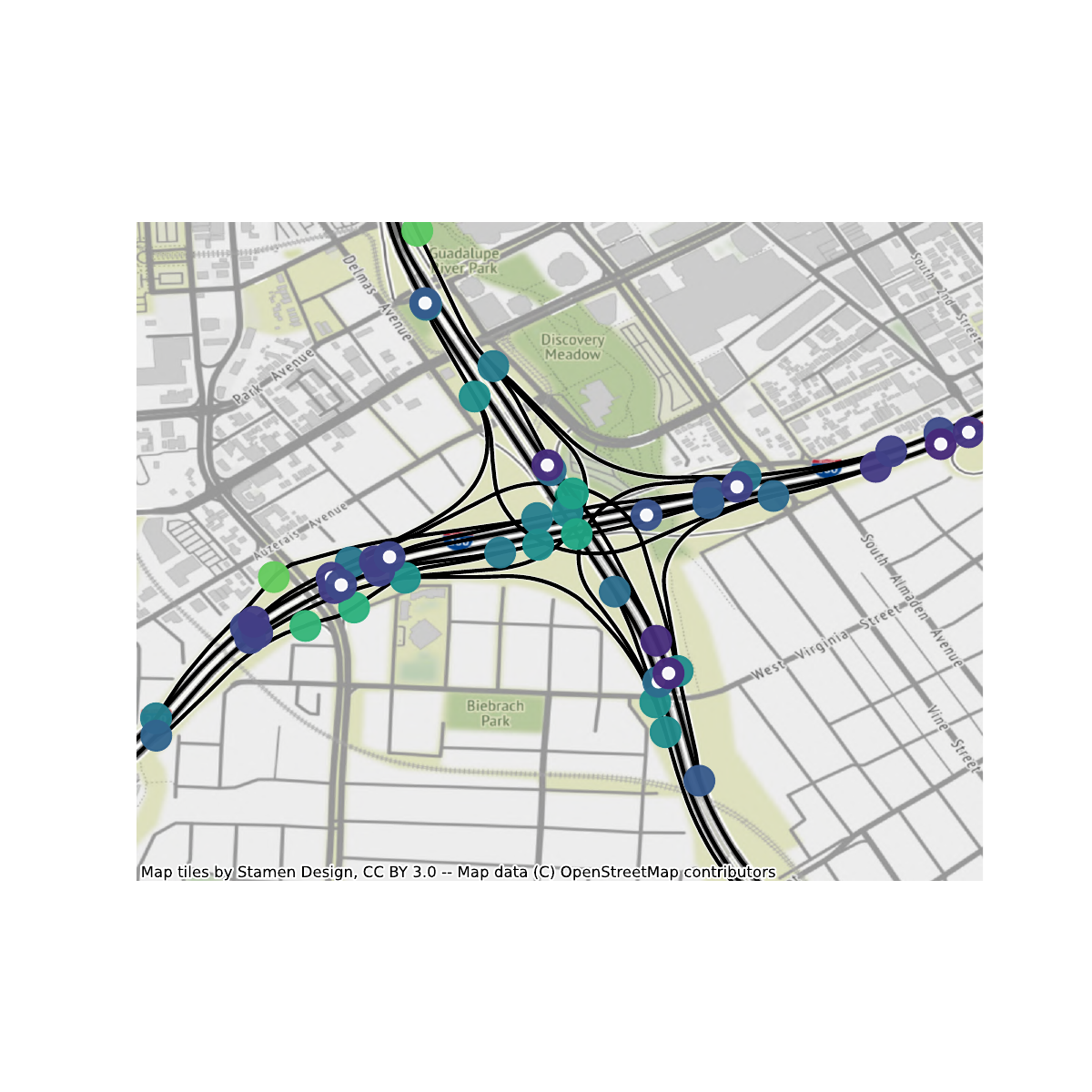}}    
        \caption{Standard deviation}
    \end{subfigure}
\caption{Traffic flow speed (mph) interpolation over a particular road junction on a graph of San Jose highways. Colored circles represent graph nodes, black lines represent edges and colored circles with a white point in the middle represent training data.}
\label{fig:traffic_junction}
\end{figure}

\paragraph{Non-conjugate learning via doubly stochastic variational inference.}

\begin{figure*}[t]
    \centering
    \begin{subfigure}[b]{0.47\textwidth}
        \centering
        \adjustbox{trim={0} {0} {0} {0},clip,width=0.99\linewidth}%
        {\includegraphics{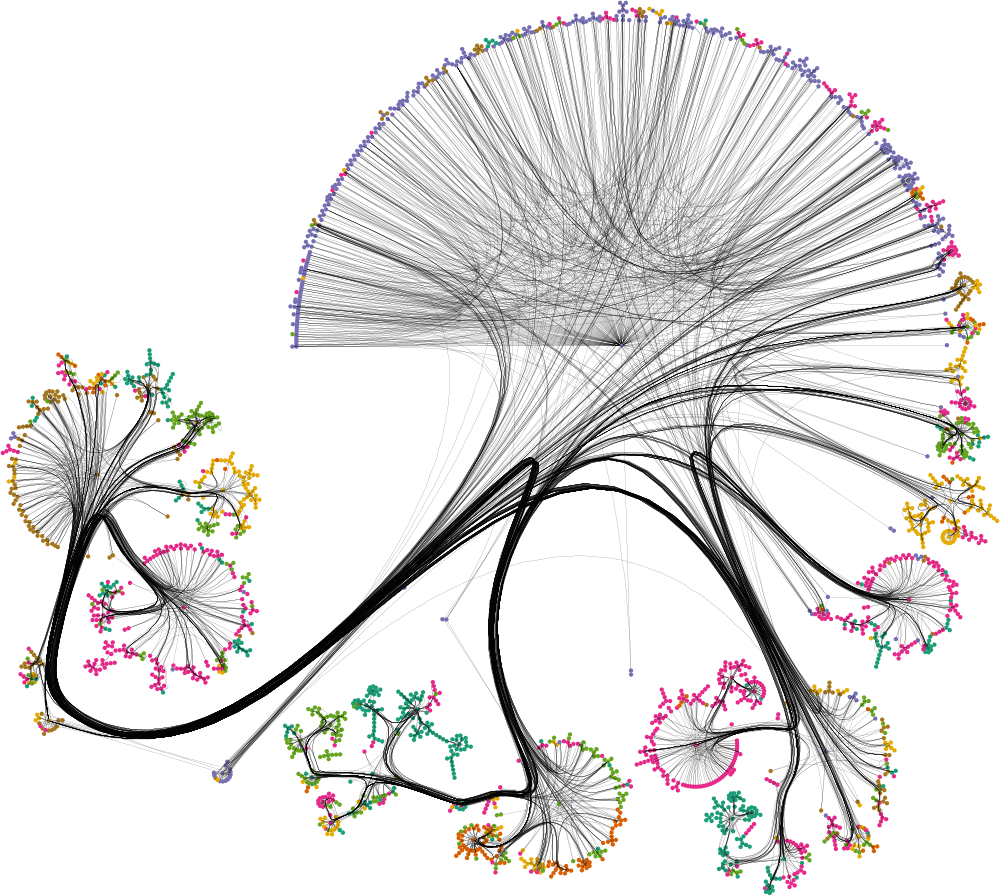}}
        \caption{Ground truth}
    \end{subfigure}
    \hspace{0.04\textwidth}
    \begin{subfigure}[b]{0.47\textwidth}
        \centering
        \adjustbox{trim={0} {0} {0} {0},clip,width=0.99\linewidth}%
        {\includegraphics{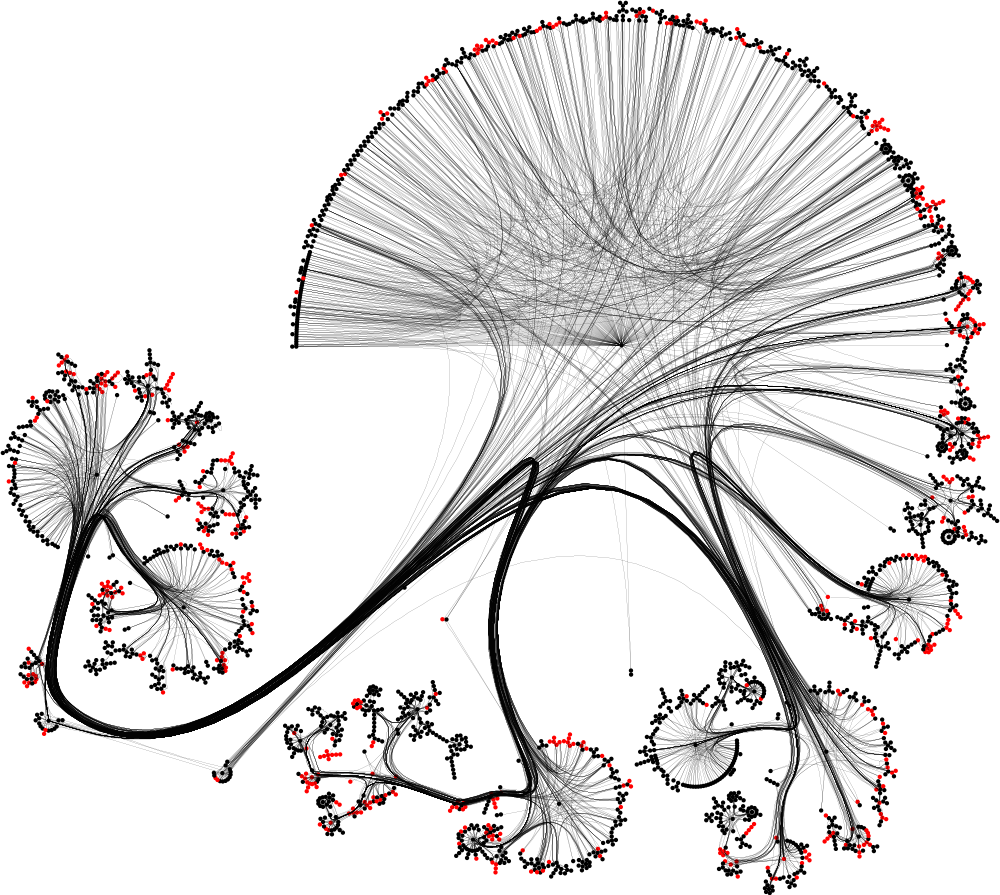}}
        \caption{Prediction errors. Red nodes are misclassified.}
    \end{subfigure}
    \caption{Topic classification for the \emph{Cora} citation network, using the citation graph.}
\label{fig:citation}
\end{figure*}

In non-conjugate settings, such as classification, the posterior distribution is no longer Gaussian, and the training techniques from the preceding sections do not apply.
Here we briefly sketch how to obtain analogous techniques using variational inference and inducing points.

Let $\v{z} \subset V$ be a set of variational \emph{inducing inputs} \cite{hensman13}, let $\v{x} \subset V$ be the training locations, let $\v{u} = \v{f}(\v{z})$.
We approximate the distribution $p(\v{u}, f \given \v{y})$ with a variational family $q(\v{u}, f)$ of the form $q(\v{u}, f) = p(f \given \v{u}) q(\v{u})$, and train the model by minimizing the Kullback--Leibler divergence
\[
&D_{\f{KL}}(q(\v{u}, f) \from p(\v{u}, f \given \v{y}))
=
\mathop{\E}_{q(\v{u},f)}\!\ln \frac{q(\v{u},f)}{p(\v{u},f\given \v{y})}
\\
&= D_{\f{KL}}(q(\v{u}) \from p(\v{u})) - \sum_{i=1}^N \mathop{\E}_{q(\v{u},f)} \ln p(y_i \given f_i)
\\
&\quad - \ln p(\v{y})\nonumber
\]
where we henceforth ignore the term $\ln p(\v{y})$ because it does not depend on $q$.

Following \textcite{titsias09a}, we choose $q(\v{u})=\f{N}(\v\mu,\m\Sigma)$ to be a free-form multivariate Gaussian.
Minimizing the Kullback--Leibler divergence therefore entails finding the closest \emph{GP posterior} to the true (non-Gaussian) posterior.
The inducing points $\v{z}$ can be chosen to be a subset of the data, or even the entire graph, provided one works with sparse precision matrices.

To minimize the Kullback--Leibler divergence, one employs a stochastic optimization algorithm, such as ADAM.
At each step, this entails sampling (1) a mini-batch of data, (2) the variational posterior from $q(\v{u},f)$ so that one can calculate the expression inside the expectation.
For classification, this expression reduces to a cross-entropy loss.
This yields the correct loss in expectation, thereby defining a doubly stochastic variational inference \cite{titsias2014} algorithm for scalable training.

\paragraph{Variational inference with interdomain inducing variables.}
An alternative to using $\v{u} = \v{f}(\v{z})$ as inducing variables is to consider \emph{interdomain inducing variables} such as $u_i = \langle f, \psi_i \rangle$. Provided that the $\psi_i$ are chosen wisely, this approach can yield orders of magnitude speed ups \cite{hensman17}. For the problem at hand, it is natural to use the kernel eigenvectors as inducing functions $\phi_i$ \cite{burt2020variational}. It is worth noting that this construction is exactly the discrete counterpart to the \emph{variational inference with spherical harmonics} method developed by \textcite{dutordoir2020} in the continuous case.

\section{Experiments}

This section is dedicated to examples that illustrate the relevance of the proposed approach. 
Full experimental details are provided in Appendix \ref{apdx:experiments}.

\subsection{Probabilistic graph interpolation of traffic data}

To illustrate the flexibility of graph GPs in settings where other GP models may not be easy to apply, we consider probabilistic interpolation of traffic congestion data of highways in the city of San Jose, California.
The graph of the road network is obtained from OpenStreetMap (OSM, \citeyear{OpenStreetMap}), and it is  pre-processed to associate to each edge a weight equal to its inverse weight and by removing hanging nodes.
We obtain traffic congestion data from the \emph{California Performance Measurement System} database \cite{chen2001}, which provides traffic flow speed in miles per hour at different locations, which we add as additional nodes onto the graph.
Note that both travel directions on a highway are considered separate roads and thus form separate edges---this can be seen in Figure~\ref{fig:traffic_junction}.

This process yields a graph with $1,016$ nodes and $1,173$ edges. Traffic flow speed is available at $325$ nodes.
We use $250$ of these nodes, chosen at random, as training data, and the remainder as test data.
The data is normalized by subtracting the mean over the training set and dividing the result by its standard deviation.
For output values, we examine traffic congestion on Monday at 17:30 in the afternoon.

We train the Mat\'{e}rn graph GP model, whose kernel is approximated using $500$ eigenpairs of the graph Laplacian.
During training, we optimize the kernel hyperparameters $\kappa$ (length scale), $\sigma^2$ (variance) and $\nu$ (smoothness), as well as the Gaussian likelihood error variance~$\varsigma^2$.

Figure \ref{fig:traffic_global} shows the predicted mean traffic on the entire graph, along with the standard deviation.
This illustrates that global dependencies are captured by the model in different regions of the graph.
In particular, the posterior GP's standard deviation increases for nodes for which traffic sensors are further away.

To illustrate local dependencies, Figure \ref{fig:traffic_junction} details the model prediction at a particular road junction. It can be seen that despite being very close in the Euclidean distance, two locations on opposite sides of the road can have associated predictions that differ considerably.
This shows that the graph GP incorporates the structure of the graph into its predictions, and allows them to differ when the graph distance is large.

\subsection{Multi-class classification in a scientific citation network}

To illustrate graph GPs in a more complex non-conjugate setting, we consider multi-class classification on the largest connected component of the \emph{Cora} citation network, considered purely as a graph, i.e. without abstracts.
This yields a data set consisting of $2,485$ nodes corresponding to scientific publications, with $5,069$ edges corresponding to citations. 
We choose $140$ nodes at random for training.
Each node is labeled according to one of seven classes, which are different scientific topics.
We employ only the graph and labels as part of the classification problem.

For the model, we use a multi-output Mat\'{e}rn graph GP where each output dimension is given using a separable kernel, yielding a GP $f : V \-> \R^7$.
We employ a categorical likelihood, with class probabilities given by applying a robust max function (Appendix \ref{apdx:experiments}) to the GP.
We approximate the graph kernel using $500$ eigenpairs of the graph Laplacian, in order to allow its pointwise evaluation, and train the model using variational inference in GPflow \cite{GPflow2017,GPflow2020multioutput}.

Performance is illustrated in Figure \ref{fig:citation}.
The fraction of misclassified nodes is small, showing that in this data set, a paper's topic can be accurately predicted from its works cited alone even in a low-data~regime.

\subsection{Comparison with other kernels}

\begin{table*}[t]
\centering
\begin{tabular}{@{}r c c c c c@{}} \toprule
 & Lap. & Mat\'{e}rn kernel & Diffusion kernel & Random walk kern. & Inverse cosine kern. \\
\midrule
\multirow{2}{20ex}{Traffic RMSE} & $\m\lap$ & 11.86 (0.69) & 11.95 (0.69) & --- & --- \\
& $\m{\widetilde{\lap}}$ & 12.82 (0.95) & 12.90 (0.95) & 13.16 (1.04) & 16.63 (0.95) \\
\addlinespace
\multirow{2}{20ex}{Citation accuracy} & $\m\lap$ & 0.79 (0.01) & 0.79 (0.02) & --- & ---\\ 
& $\m{\widetilde{\lap}}$ & 0.79 (0.01) & 0.78 (0.01) & 0.78 (0.02) & 0.48 (0.02)\\ 
\bottomrule
\end{tabular}
\caption{Mean and standard deviation (in parentheses) of model performance for different graph GP kernels.
Note that the standard deviation of the ground truth traffic flow speed across the entire graph is approximately $17.1$~mph. We denote the unnormalized Laplacian by $\m\lap$, and the symmetric normalized Laplacian by $\m{\widetilde{\lap}} = \m{D}^{-1/2} \m\lap \m{D}^{-1/2}$.
}
\label{table:mat_diff_comp}
\end{table*}

We now evaluate the performance of GPs induced by the graph Mat\'{e}rn on both of the examples considered.
For comparison, we consider the graph diffusion kernel, which is the limit of the Mat\'{e}rn kernel as $\nu\->\infty$, as well as the random walk and inverse cosine kernels defined in \textcite{smola2003}.
We consider variants employing both normalized and unnormalized graph Laplacians.
We evaluate performance via mean squared error, using $250$ training nodes with $75$ test nodes for the traffic example, and $140$ training nodes with $1000$ test nodes for the citation network example.
We repeat this $10$ times to obtain average mean squared error and accuracy.
Results are presented in Table \ref{table:mat_diff_comp}.
This shows that use of the more flexible Mat\'{e}rn class results in similar or slightly improved performance compared to alternatives on these data sets.

\section{Conclusion}

In this work, we study graph Mat\'{e}rn Gaussian processes, which are defined as analogous to the Mat\'{e}rn stochastic partial differential equations used in the Euclidean and Riemannian cases.
We discuss a number of their properties, and provide scalable training algorithms by means of Gaussian Markov random fields, graph Fourier features and sparse GPs.
We demonstrate their effectiveness on a simple graph interpolation problem, and on a multi-class graph classification problem.
Compared to other graph kernels, the graph Mat\'{e}rn class is flexible, interpretable, and shown to perform well, mirroring the behavior of Euclidean and Riemannian Mat\'{e}rn kernels.
We hope these techniques inspire new use of Gaussian processes by machine learning practitioners.

\section*{Acknowledgments}

We are grateful to Mohammed Charrout for finding a bug in our software, which enabled us to fix incorrect results in Table \ref{table:mat_diff_comp} that were present in prior drafts of this manuscript.
This research was partially supported by "Native towns", a social investment program of PJSC Gazprom Neft and by the Ministry of Science and Higher Education of the Russian Federation, agreements N\textsuperscript{\underline{o}} 075-15-2019-1619 and N\textsuperscript{\underline{o}} 075-15-2019-1620.

\section*{Erratum}

In the original published manuscript, Table~\ref{table:mat_diff_comp} incorrectly labeled RMSE values as MSE.
In addition, the RMSE values presented were incorrectly scaled by a constant, and were equal to $\frac{1}{\sqrt{75}}\f{RMSE}$ instead of $\f{RMSE}$.
This scaling difference was applied to both the proposed method and baselines, and does not affect relative comparisons between methods.
The current manuscript version corrects both aforementioned errors.

\printbibliography

\onecolumn
\appendix

\section{Appendix: experimental details}
\label{apdx:experiments}

\subsection*{Probabilistic graph interpolation of traffic data}

Here we consider the problem of interpolating traffic congestion data over a road network, consisting of San Jose highways.
We build the road network, a weighted graph, from OpenStreetMap data \cite{OpenStreetMap} obtained using the OSMnx library \cite{boeing2017}.

We obtain traffic congestion data from the \emph{California Performance Measurement Systems} database \cite{chen2001}.
This system maps sensor location and time \& date to (among other factors) traffic flow speed in miles per hour measured by the given sensor at a particular time and date.
We chose the specific date and time to be 17:30 on Monday, the 2\textsuperscript{th} of January, 2017.
We bind the traffic congestion data to the graph by adding additional nodes that subdivide existing edges of the graph at the location of the measurement points.
The edge weights are assigned to be equal to the inverse travel distances between the corresponding nodes.
When we encounter two nodes connected by multiple edges, we remove all but one edge between them, and set its weight to be the inverse mean travel distance between the pair of nodes. 
After this, the graph is processed by (1) twice subsequently removing hanging nodes, and (2) removing two other nodes located particularly far away from data.
We end up with $325$ labeled nodes on a sparse connected graph with $1016$ nodes and $1173$ edges.

For the model, we employ a GP with a graph Mat\'{e}rn kernel and the graph diffusion kernel.
Both kernels are approximated using $500$ eigenpairs of the graph Laplacian.
The eigenpairs required to build graph kernels are computed using TensorFlow's eigenvalue decomposition \textsc{(tf.linalg.eigh)} routine and are thus exact up to a numerical error arising from the floating point arithmetic.
We train by optimzing the kernel hyperparameters $\kappa$ (length scale), $\sigma^2$ (variance) and for the case of the graph Mat\'{e}rn kernel, $\nu$ (smoothness), as well as the Gaussian likelihood variance $\varsigma^2$.
We use the following values for initialization: $\nu = 3/2$ for the graph Mat\'{e}rn kernel, $\kappa = 3$, $\sigma^2 = 1$, $\varsigma^2 = 0.01$.
We randomly choose $250$ nodes with traffic flow rates as training data.
The likelihood is optimized for $20000$ iterations with the ADAM optimizer, with learning rate set to the TensorFlow default value of $0.001$.

We repeat the experiment 10 times.
The average RMSE over the 10 experiments and its standard deviation are presented in Table \ref{table:mat_diff_comp} for the graph Mat\'{e}rn kernel and for the graph diffusion kernel.
The smoothness parameter $\nu$ after optimization varies across the runs.
Its mean is $1.8$ and the standard deviation is $0.9$.
One of these ten runs is visualized in Figures \ref{fig:traffic_global} and \ref{fig:traffic_junction}.

\subsection*{Multi-class classification in a scientific citation network}

Here we consider multi-class classification on the largest connected component of the \emph{Cora} citation network.

There are $2708$ nodes corresponding to scientific publications in the \emph{Cora} citation network.
Usually, each node of this graph is attributed by a $0$/$1$-valued word vector indicating the absence/presence of the corresponding word from the dictionary of $1433$ unique words in the abstract---we discard this portion of data to focus on other problem aspects.\footnote{It's also possible to use a separable kernel given by the product of a graph Mat\'{e}rn kernel and a Euclidean kernel operating on word vectors, but this model is unlikely to perform well due to usual GP difficulties in high dimension. Incorporating this portion of the data effectively therefore necessitates additional modeling considerations, so we do not focus on it here.}
Two nodes are connected by an edge if one paper cites another. 
The edges are unweighted and undirected, and there are $5429$ edges overall.
The nodes are labeled into one of seven classes.
We concentrate on the largest connected component of the Cora graph, which has $2485$ nodes and $5069$ edges.

We consider a Gaussian process $V \-> \R^7$ with independent components as a latent process for the classification task.
The process is then combined with the robust max function $\v\varphi(\cdot)$ to yield a vector of probabilities over the classes.
Let $\v{f} = (f_1, .., f_7)$ denote the values of $7$-dimensional latent process. Then we have
\[
\v\varphi(\v{f}) &= (\varphi_1(\v{f}), .., \varphi_7(\v{f}))
&
\varphi_c(\v{f}) &=
\begin{cases}
1-\varepsilon, &\t{if} c = \argmax_c f_c \\
\varepsilon/6, &\t{otherwise},\\
\end{cases}
\]
where $\varepsilon$ is fixed to be $10^{-3}$.
For the likelihood, we employ the categorical distribution.
We use the graph Mat\'{e}rn kernel and the graph diffusion kernel for the prior. 
Both kernels are approximated using $500$ eigenpairs of the graph Laplacian.
The eigenpairs required are computed using TensorFlow eigenvalue decomposition \textsc{(tf.linalg.eigh)} and are thus exact up to a numerical error arising from use of floating point arithmetic.
We use $140$ random nodes as a training set and predict labels for the remaining $2345$ nodes.

For the variational GP, we employ the standard SVGP model of GPflow \cite{GPflow2017,GPflow2020multioutput}. We fix the the coordinates of inducing points, which are located on the graph, to the data locations, and do not optimize them.
We restrict the covariance of the inducing distribution to a diagonal form, since this results in the best performance in this setting.
The inducing mean is initialized to be zero and the inducing covariance is initialized to be the identity.
The prior hyperparameters are initialized to be
$\nu = 3$ for the Mat\'{e}rn kernel, $\kappa = 5$, $\sigma^2 = 1$.
The variational objective is optimized for $20000$ iterations with the ADAM optimizer whose learning rate set to $0.001$. 
Convergence is usually achieved in a significantly lower number of iterations.
The SVGP batch size is equal to the number of training data points, and whitening is enabled.

\begin{figure}[t]
        \adjustbox{trim={0} {0} {0} {0},clip,width=\linewidth}%
        {\includegraphics{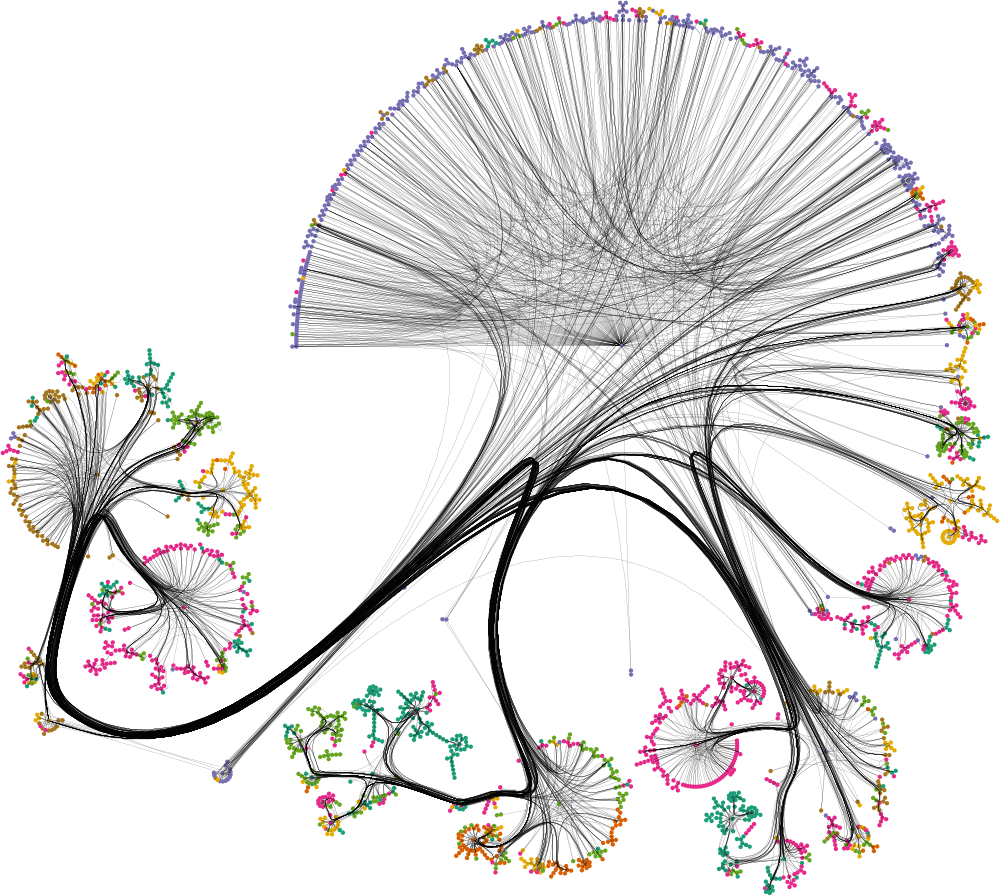}}
        \caption{Topic classification for the \emph{Cora} citation network, using the citation graph. Ground truth.}
        \label{fig:citation_gt_big}
\end{figure}

\begin{figure}[t]
        \adjustbox{trim={0} {0} {0} {0},clip,width=\linewidth}%
        {\includegraphics{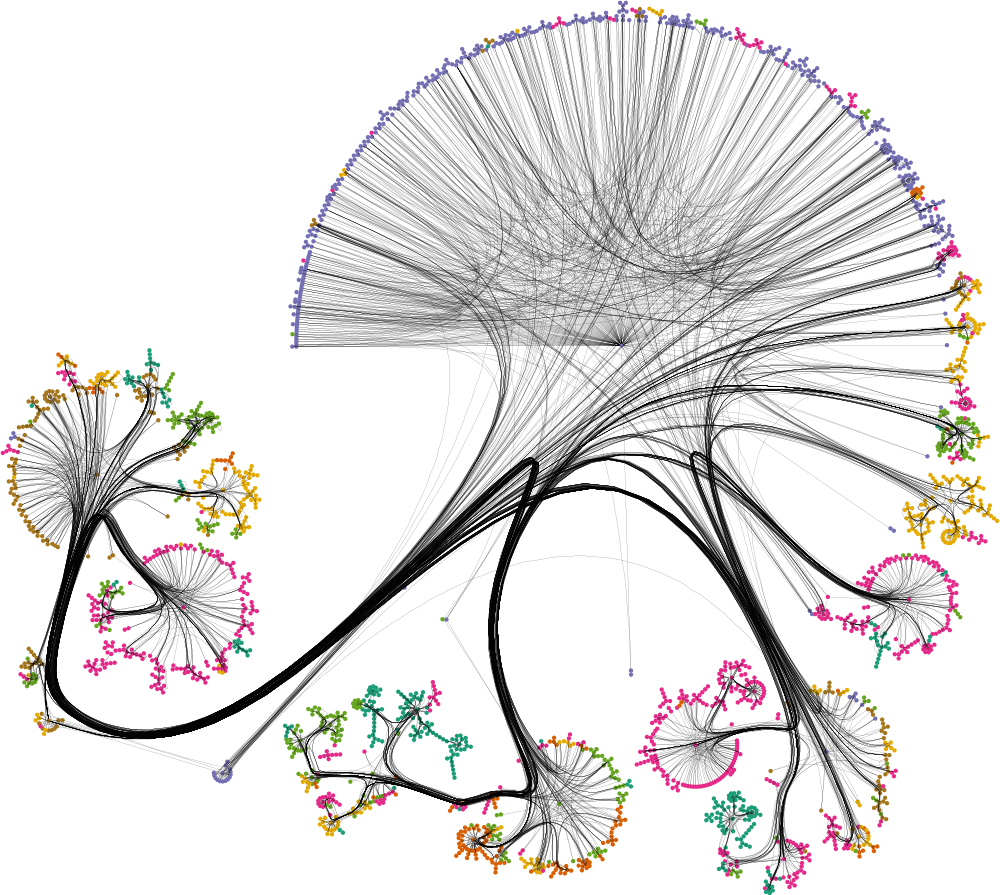}}
        \caption{Topic classification for the \emph{Cora} citation network, using the citation graph. Prediction.}
        \label{fig:citation_pred_big}
\end{figure}

\begin{figure}[t]
        \adjustbox{trim={0} {0} {0} {0},clip,width=\linewidth}%
        {\includegraphics{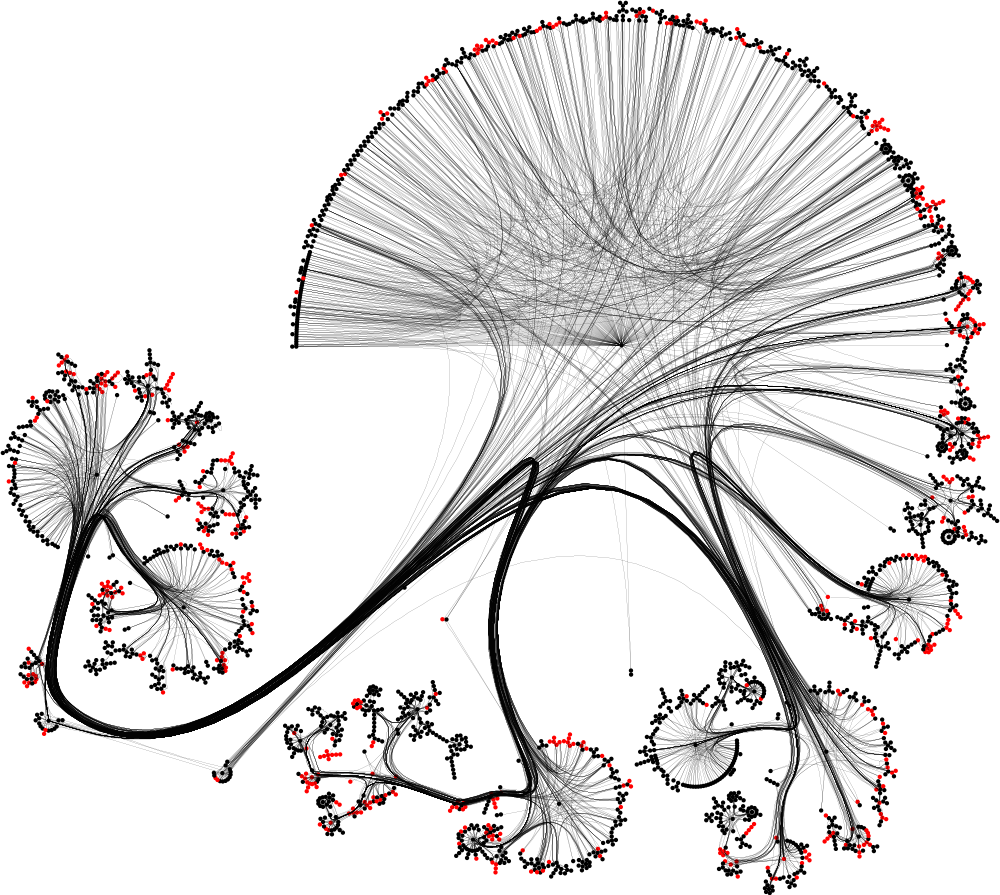}}
        \caption{Topic classification for the \emph{Cora} citation network, using the citation graph. Prediction errors. Red nodes are misclassified.}
        \label{fig:citation_error_big}
\end{figure}

We repeat the experiment 10 times.
Each time, we additionally randomly choose a test set of $1000$ nodes.
The average (over the 10 experiments) accuracy (fraction of the correctly classified points) and its standard deviation are presented in Table \ref{table:mat_diff_comp} for the graph Mat\'{e}rn kernel and for the graph diffusion kernel.
The value of the smoothness parameter $\nu$ after optimization varies considerably across runs.
The mean $\nu$ is $3.2$ and the standard deviation is $2.7$.
The results of one of said ten runs are visualized on Figures \ref{fig:citation_gt_big}, \ref{fig:citation_pred_big}, \ref{fig:citation_error_big}.
Note that Figures \ref{fig:citation_gt_big} and \ref{fig:citation_error_big} repeat Figure \ref{fig:citation} (a), (b) at a higher resolution.

\section{Appendix: graph Fourier features} \label{apdx:theory}

Assume that $\m\lap = \m{U} \m{\Lambda} \m{U}^T$ with $\m{U}$ orthogonal and $\m{\Lambda}$ diagonal.
For a set of of locations $\v{x} \in V^n$ define $\m{U}(\v{x})$ to be the $n \x d$ submatrix of $\m{U}$ with rows corresponding to the elements of $\v{x}$.
Define also $\Psi(\lambda) = \Phi(\lambda)^{-2}$, where $\Phi$ is given in \eqref{eqn:Phi_def}.
Then, if $f$ is a graph Mat\'{e}rn process and $\m{K}_{\v{x} \v{x}}$ is the covariance matrix of $f(\v{x})$, we have
\[
\m{K}_{\v{x} \v{x}}
=
\m{U}(\v{x})
\Psi(\m{\Lambda})
\m{U}(\v{x})^T
\approx
\widetilde{\m{U}}(\v{x})
\Psi(\widetilde{\m{\Lambda}})
\widetilde{\m{U}}(\v{x})^T
,
\]
where $\widetilde{\m{U}}(\v{x})$ and $\widetilde{\m{\Lambda}}$ are the approximations of $\widetilde{\m{U}}(\v{x})$ and $\widetilde{\m{\Lambda}}$ with elements in the last $d-l$ rows of matrices $\widetilde{\m{U}}(\v{x})$ and $\widetilde{\m{\Lambda}}$ set to zero.
This approximation arises from finding only the $l$ smallest eigenpairs, which can be done more efficiently than the full eigendecomposition by employing, for instance, the Lanczos algorithm. 

With this, using the Woodbury matrix identity, we get
\[
\del{\m{K}_{\v{x} \v{x}} + \sigma^2 \m{I}}^{-1}
&=
\sigma^{-2} \m{I}
-
\sigma^{-4}
\m{U}(\v{x})
\del[1]{\Psi(\m{\Lambda})^{-1} + \overbracket[0.1ex]{\sigma^{-2} \m{U}(\v{x})^T \m{U}(\v{x})}^{\sigma^{-2} \m{I}}}^{-1}
\m{U}(\v{x})^T
,
\\
\del{\m{K}_{\v{x} \v{x}} + \sigma^2 \m{I}}^{-1}
&\approx
\sigma^{-2} \m{I}
-
\sigma^{-4}
\widetilde{\m{U}}(\v{x})
\del[1]{\Psi(\widetilde{\m{\Lambda}})^{-1} + \underbracket[0.1ex]{\sigma^{-2} \widetilde{\m{U}}(\v{x})^T \widetilde{\m{U}}(\v{x})}_{\text{diagonal}}}^{-1}
\widetilde{\m{U}}(\v{x})^T
,
\]
where a diagonal matrix is inside the brackets.
It is clear that computing $\del{\m{K}_{\v{x} \v{x}} + \sigma^2 \m{I}}^{-1} \v{y}$ when the (approximation of) eigenpairs are known can be done without incurring the $O(n^3)$ cost, resulting in efficient computation.

The resulting kernel can be (approximately) expressed by
\[
k(i, j)
=
\sum_{s=1}^d
\Psi(\lambda_s) \v{u_s}(i) \v{u_s}(j)
\approx
\sum_{s=1}^l
\Psi(\lambda_s) \v{u_s}(i) \v{u_s}(j),
\]
where $\lambda_s$ are eigenvalues of $\m\lap$, while $\v{u_s}(i)$ and $\v{u_s}(j)$ are the $i$-th and $j$-th component of the eigenvector $\v{u_s}$ corresponding to $\lambda_s$.
For Mat\'{e}rn kernels we have $\Psi(\lambda) = \del{\frac{2 \nu}{\kappa^2} + \lambda}^{-\nu}$ from which it is clear that differentiating $k$ with respect to the smoothness parameter $\nu$ is straightforward.
This allows to optimize the marginal likelihood over the smoothness parameter $\nu$ in the same way this optimization over the length scale parameter $\kappa$ is usually done.

\end{document}